\documentclass[10pt,twocolumn,letterpaper]{article}

\usepackage[pagenumbers]{cvpr} 

\usepackage[dvipsnames]{xcolor}

\definecolor{cvprblue}{rgb}{0.21,0.49,0.74}
\usepackage[pagebackref,breaklinks,colorlinks,citecolor=cvprblue]{hyperref}
\usepackage[dvipsnames]{xcolor}
\usepackage{tabularx}
\definecolor{Background}{HTML}{FFFFFF}

\def\paperID{*****} 
\def\confName{CVPR}
\def\confYear{2024}

\title{\colorbox{Background}{\textcolor{Orange}{DocXChain:}}~A Powerful Open-Source Toolchain for Document Parsing and Beyond}

\author{Cong Yao\\
Alibaba DAMO Academy\\
Beijing, China\\
{\tt\small Correspondence to:~yaocong2010@gmail.com}
}

\begin{document}
\maketitle

\begin{abstract} 
In this report, we introduce \textbf{DocXChain}, a powerful open-source toolchain for document parsing, which is designed and developed to automatically convert the rich information embodied in \textbf{unstructured documents}, such as text, tables and charts, into \textbf{structured representations} that are readable and manipulable by machines. Specifically, basic capabilities, including text detection, text recognition, table structure recognition and layout analysis, are provided. Upon these basic capabilities, we also build a set of fully functional pipelines for document parsing, i.e., general text reading, table parsing, and document structurization, to drive various applications related to documents in real-world scenarios. Moreover, DocXChain is concise, modularized and flexible, such that it can be readily integrated with existing tools, libraries or models (such as LangChain and ChatGPT), to construct more powerful systems that can accomplish more complicated and challenging tasks. The code of DocXChain is publicly available at:~\url{https://github.com/AlibabaResearch/AdvancedLiterateMachinery/tree/main/Applications/DocXChain}
\end{abstract}

\section{Introduction} \label{sec:Introduction}

\textit{\footnotesize ``Make Every Unstructured Document Literally Accessible to Machines''}
\vspace{-2mm}
\begin{flushright}
{\footnotesize \textbf{-- The DocXChain Development Team, 2023}}
\end{flushright}

Documents are ubiquitous\footnote{In this project, we adopt the \textbf{broad concept of documents}, meaning that DocXChain can support various kinds of documents, including regular documents (such as books, academic papers and business forms), street view photos, presentations and even screenshots.}, since they are excellent carriers for recording and spreading information across space and time. Documents have been playing a critically important role in the daily work, study and life of people all over the world. Every day, billions of documents in different forms are created, viewed, processed, transmited and stored around the world, either physically or digitally. However, not all documents in the digital world can be directly accessed by machines (including computers and other automatic equipments), as only a portion of the documents can be successfully parsed with low-level procedures. For instance, the Adobe Extract APIs are able to directly convert the metadata of born-digital PDF files into HTML-like trees~\cite{SaadFalcon2023PDFTriageQA}, but would completely fail when handling PDFs generated from photographs produced by scanners or images captured by cameras. Therefore, if one would like to make documents that are not born-digital conveniently and instantly accessible to machines, a powerful toolset for extracting the structures and contents from such unstructured documents~\cite{Zhu2015SceneTD, Long2018SceneTD, Cui2021DocumentAB} is of the essence.

In this article, we introduce a new open-source toolchain for document parsing, called DocXChain, which is dedicated to converting unstructured documents into structured representations. Concretely, DocXChain provides tools to precisely detect layouts, read text and extract tables of documents, and arrange these elements in an organized manner, such that the rich and precious information embodied in various unstructured documents, which is previously not accessible to machines, has been unlocked, and a mass of applications related to documents are henceforth possible.

DocXChain is unique and powerful in that: (1) It assembles a collection of industry-leading algorithmic models for text detection, text recognition, table structure recognition and layout analysis, which are open-sourced by our team and publicly available on ModelScope\footnote{\url{https://github.com/modelscope/modelscope}} and AdvancedLiterateMachinery\footnote{\url{https://github.com/AlibabaResearch/AdvancedLiterateMachinery}}; (2) Different from existing open-source libraries for OCR and document parsing, the tools in DocXChain can effectively handle documents from real-world scenarios, in addition to those collected for pure academic purposes; (3) DocXChain works out-of-the-box and is compatible with other tools or models (\eg, LangChain~\cite{LangChain} and ChatGPT~\cite{ChatGPT}), since it is concise and modularized.

\section{Design and Implementation of DocXChain} \label{sec:DocXChain}

In this section, we will describe in detail the design and implementation of DocXChain.

\subsection{Core Ideology}  \label{sec:CoreIdeology}

The core design ideas of DocXChain are three-fold:
\begin{itemize}
  \item \textbf{Object:} The central objects of DocXChain are \textbf{\textit{documents}}, rather than \textbf{\textit{LLMs}}.
  \item \textbf{Concision:} The capabilities for document parsing are presented in a simple ``\texttt{modules} + \texttt{pipelines}'' fashion, while unnecessary abstraction and encapsulation are abandoned.
  \item \textbf{Compatibility:} This toolchain can be used as a stand-alone procedure to structurize documents, while it can also be readily integrated with existing tools, libraries or models, such as LangChain~\cite{LangChain}, ChatGPT~\cite{ChatGPT} and GPT-4~\cite{GPT-4}, to build more powerful systems that can solve more complicated and challenging tasks.
\end{itemize}

\subsection{System Overview}  \label{sec:SchematicOverview}

\begin{figure}[ht]
  \centering
  \vspace{-3mm}
  \includegraphics[width=0.48\textwidth]{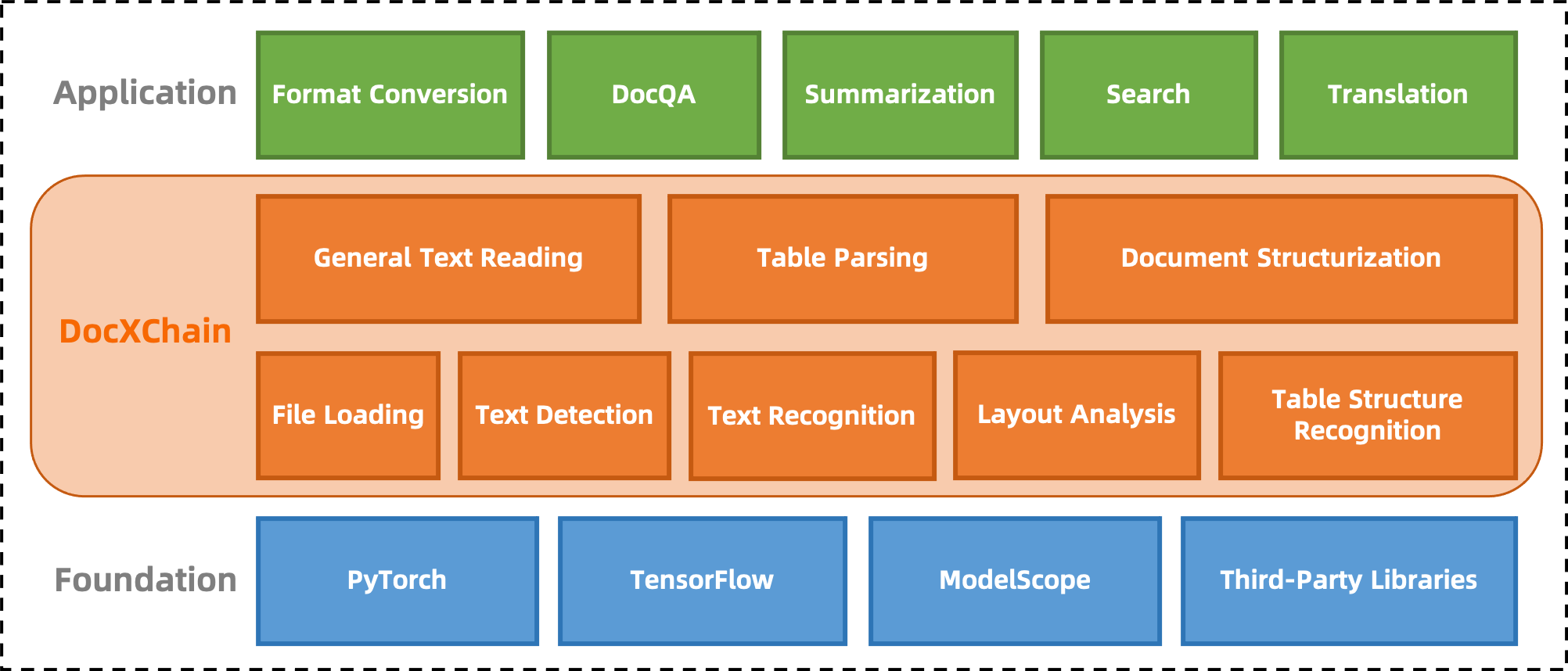}
  \vspace{-5mm}
  \caption{System overview of DocXChain.}
  \label{fig:overview}
  \vspace{-1mm}
\end{figure}

The overview of DocXChain is illustrated in Fig.~\ref{fig:overview}. DocXChain provides atomic capabilities as well as fully functional pipelines, which are built upon PyTorch~\cite{Paszke2019PyTorchAI}, TensorFlow~\cite{Abadi2016TensorFlowAS}, ModelScope~\cite{ModelScope} and other 3rd-party libraries (such as the libraries for loading images and PDFs).

In general, DocXChain, as a middle-level tool set, can be adopted to support high-level applications related to documents, such as document format conversion (\eg, pdf2word and image2word), DocQA, summarization, search and translation~\cite{Cui2021DocumentAB}.

\subsection{Modules and Pipelines}  \label{sec:ModulesPipelines}

\iftrue
\begin{table}[htp]
  \centering
  \begin{tabularx}{0.5\textwidth}{l X}
    \toprule
    \textbf{Module} & \textbf{Function Description} \\
    \midrule
    \scriptsize{\texttt{File Loading}} & \small{Load document files. Only images (.jpg and .png) and PDFs (.pdf) are supported currently.} \\
    \scriptsize{\texttt{Text Detection}} & \small{Detect all text instances (those virtually machine-identifiable).} \\
    \scriptsize{\texttt{Text Recognition}} & \small{Recognize each text instance (assume that text detection has been perfomed in advance).} \\
    \scriptsize{\texttt{Layout Analysis}} & \small{Identify and categorize all layout regions (those virtually machine-identifiable).} \\
    \scriptsize{\texttt{Table Structure Recognition}} & \small{Recognize the structure of the given table. At present, only tables with \textbf{visible borders} are supported.} \\
    \bottomrule
  \end{tabularx}
  \vspace{-2mm}
  \caption{Function description of the modules in DocXChain.}
  \label{tab:Modules}
\end{table}
\fi

The detailed descriptions of the the basic modules in DocXChain are depicted in Tab.~\ref{tab:Modules}. Each basic module realizes an atomic capability. DocXChain accepts image and PDF\footnote{PDF pages will be converted to images before subsequent processing. By default, only the first page will be chosen and parsed if the input PDF file has multiple pages.} files as input. Currently, the supported languages are Chinese and English.

\begin{table}[htp]
  \centering
  \begin{tabularx}{0.5\textwidth}{l X}
    \toprule
    \textbf{Pipeline} & \textbf{Function Description} \\
    \midrule
    \scriptsize{\texttt{General Text Reading}} & \small{Detect and recognize all text instances (those virtually machine-identifiable).} \\
    \scriptsize{\texttt{Table Parsing}} & \small{Perform table parsing (table structure recognition + textual content recognition).} \\
    \scriptsize{\texttt{Document Structurization}} & \small{Structurize the given document (layout analysis + text detection and recognition).} \\
    \bottomrule
  \end{tabularx}
  \vspace{-2mm}
  \caption{Function description of the pipelines in DocXChain.}
  \label{tab:Pipelines}
  \vspace{-3mm}
\end{table}

The detailed descriptions of the pipelines in DocXChain are shown in Tab.~\ref{tab:Modules}. These typical pipelines are built with the basic modules in DocXChain. For example, the \textbf{General Text Reading} pipeline consists of the \textbf{Text Detection} module and the \textbf{Text Recognition} module. For certain, one could make more pipelines to meet different requirements with the modules of DocXChain and other tools or libraries.

\subsection{Qualitative Examples}  \label{sec:QualitativeExamples}

\begin{figure}[ht]
  \centering
  \vspace{-2mm}
  \includegraphics[width=0.45\textwidth]{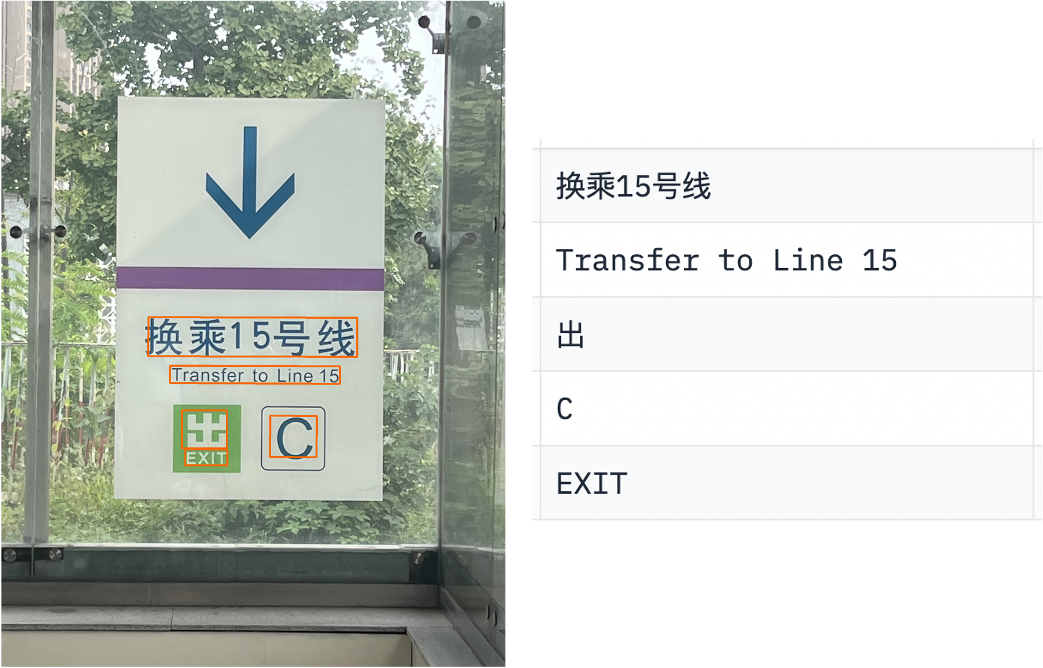}
  \vspace{-2mm}
  \caption{General text reading example. The text detections are represented with orange quadrangles, while the text contents are listed on the right panel.}
  \label{fig:text_reading_example}
\end{figure}

\begin{figure}[ht]
  \centering
  \vspace{-1mm}
  \includegraphics[width=0.48\textwidth]{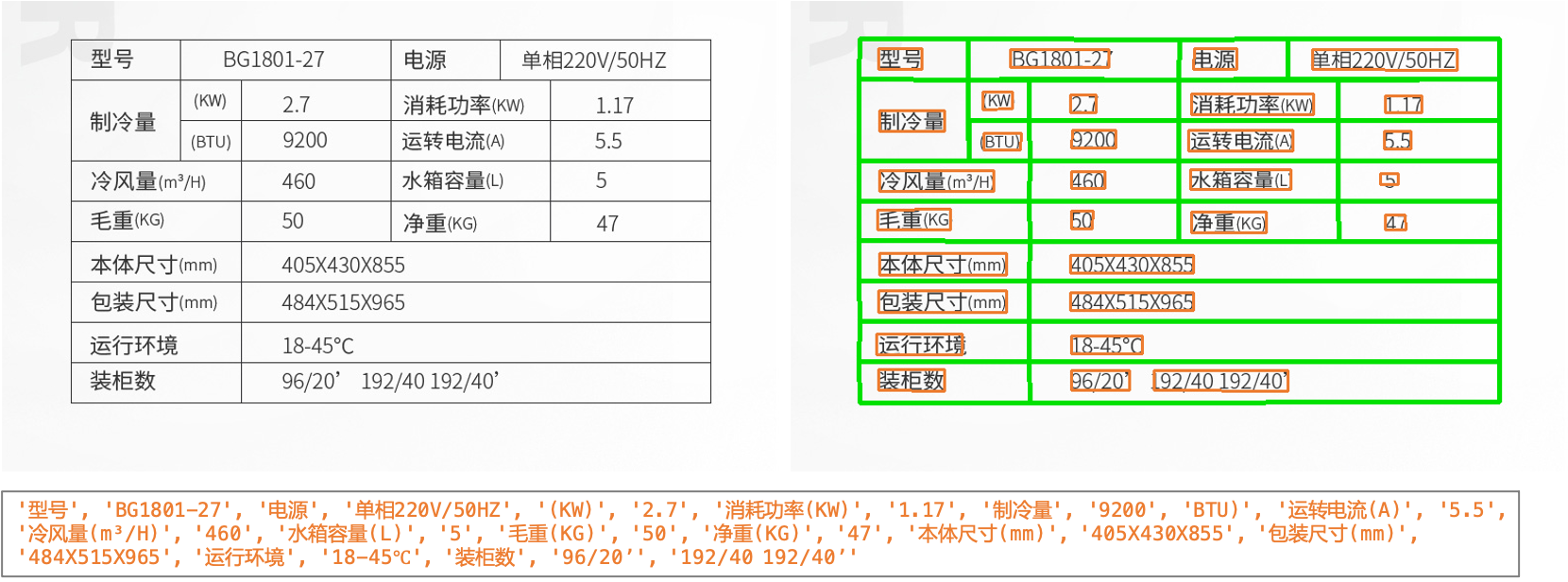}
  \vspace{-5mm}
  \caption{Table parsing example. The original image is shown on the left, while the table cells (in green) and text detections (in orange) are depicted on the right. For clarity, the recognized text contents are not overlaid on the image, but listed in the box below.}
  \label{fig:table_parsing_example}
  \vspace{-5mm}
\end{figure}

\begin{figure}[ht]
  \centering
  \vspace{-2mm}
  \includegraphics[width=0.48\textwidth]{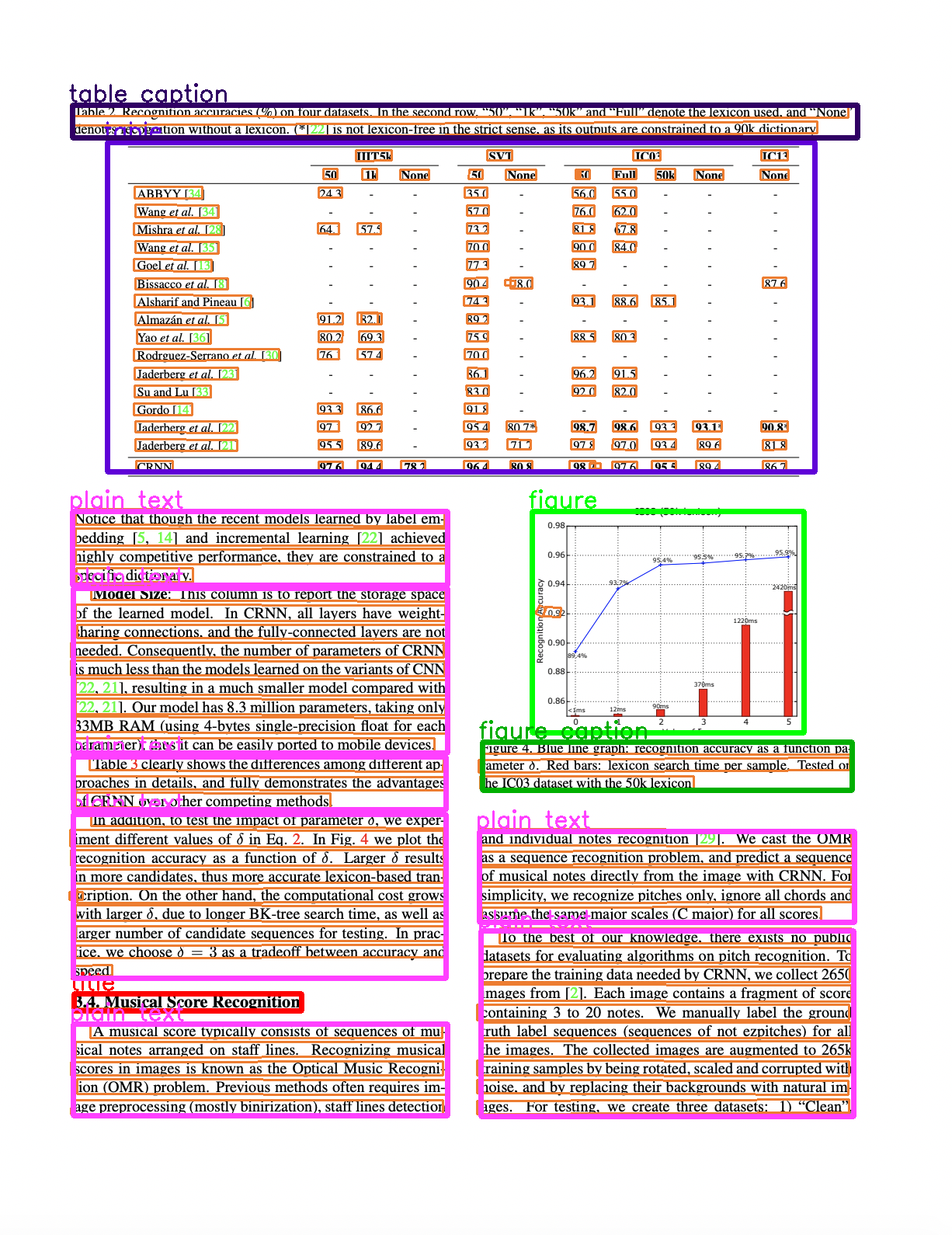}
  \vspace{-12mm}
  \caption{Document structurization example. Different colors are used to illustrate the categories of different layout regions. The text detections are represented with orange quadrangles. For clarity, the recognized text contents are skipped.}
  \label{fig:document_structurization_example}
  \vspace{-2mm}
\end{figure}

We also evaluate DocXChain on a small set of documents from real-world scenarios. As shwon in Fig.~\ref{fig:text_reading_example}, ~\ref{fig:table_parsing_example} and~\ref{fig:document_structurization_example}, DocXChain is able to successfully handle documents from different scenarios that are quite common in reality.

Specifically, it can read subway transfer information on a signboard (Fig.~\ref{fig:text_reading_example}); it is also able to extract the structure and textual contents of a table containing detailed product specifications (Fig.~\ref{fig:table_parsing_example}); for documents with complex layout and dense text, it is capable of comprehensively parsing and organizing all the key elements (Fig.~\ref{fig:document_structurization_example}). In brief, the wide adaptability and high flexibility of DocXChain makes it an excellent choice to power various real-world applications.

\section{Conclusion and Outlook} \label{sec:Conclusion}

In this article, we have introduced DocXChain, an open-source toolchain for document parsing. It releases algorithmic models and engineering codes to support basic capabilities as well as typical pipelines, which can be used to extract the structures and contents from unstructured documents.

We also notice that the newly released GPT-4V(ision)~\cite{GPT-4V(ision)} is capable of reading text from images, understanding charts and reasoning with tables. However, GPT-4V(ision) is not a open-source system, and further quantitative investigations are needed to validate its accuracy and robustness in challenging scenarios~\cite{Yang2023TheDO}. Therefore, our DocXChain, as a lightweight, open-source specialist toolchain for precise document parsing, is definitely highly complementary to such generalists, when analysing and understanding documents in real-world applications.

DocXChain is designed and developed with the original aspiration of promoting the level of digitization and structurization for documents. In the future, we will go beyond pure document parsing capabilities, to explore more possibilities, e.g., combining DocXChain with large language models (LLMs) to perform document information extraction (IE), question answering (QA) and retrieval-augmented generation (RAG).

{
    \small
    \bibliographystyle{ieeenat_fullname}
    \bibliography{DocXChain}
}


\end{document}